# Automatic Robotic Development through Collaborative Framework by Large Language Models

Zhirong Luan*
School of Electrical Engineering
Xi'an University of Technology
Xi'an, China
luanzhirong@xaut.edu.cn

Yujun Lai
School of Electrical Engineering
Xi'an University of Technology
Xi'an, China
2221920082@stu.xaut.edu.cn

Rundong Huang
School of Electrical Engineering
Xi'an University of Technology
Xi'an, China
3211712276@stu.xaut.edu.cn

Xiaruiqi Lan
College of Artificial Intelligence
Xi'an Jiaotong University
Xi'an, China
lanxiaruiqi@stu.xjtu.edu.cn

Liangjun Chen
College of Artificial Intelligence
Xi'an Jiaotong University
Xi'an, China
liangjunchen@xjtu.edu.cn

Badong Chen
College of Artificial Intelligence
Xi'an Jiaotong University
Xi'an, China
chenbd@mail.xjtu.edu.cn

*Abstract*—Despite the remarkable code generation abilities of large language models (LLMs), they still face challenges in complex task handling. Robot development, a highly intricate field, inherently demands human involvement in task allocation and collaborative teamwork[1]. To enhance robot development, we propose an innovative automated collaboration framework inspired by real-world robot developers. This framework employs multiple LLMs in distinct roles—analysts, programmers, and testers. Analysts delve deep into user requirements, enabling programmers to produce precise code, while testers fine-tune the parameters based on user feedback for practical robot application. Each LLM tackles diverse, critical tasks within the development process. Clear collaboration rules emulate real-world teamwork among LLMs. Analysts, programmers, and testers form a cohesive team overseeing strategy, code, and parameter adjustments [2]. Through this framework, we achieve complex robot development without requiring specialized knowledge, relying solely on non-experts' participation.

## I. INTRODUCTION

In reality, developing a robot is a complex and difficult task, and the difficulty lies in the decision making of the task and the ability to generate the correct and appropriate code. The large-scale language model (LLM) can be used as a tool to understand the user's needs and generate a professional robot development strategy, and then generate the code for robot control.

Code generation has drawn great attention in recent years. The successful code generation can improve the efficiency and quality of the robot development. The step of code generation is particularly important in the process of robot development. Nevertheless, generating the correct code is difficult for complex tasks on robot development. In fact, in the past, such a complex task required a team of developers with rich experience to collaborate with each other.

LLMs can be used need to enable teamwork with its different capabilities. Specifically, firstly we can train models with different capabilities to handle multi-seeded tasks; secondly we jointly train models to promote mutual understanding and finally combine different LLMs into development teams [3]. However, it is difficult to train large language models independently, which is exacerbated by the need for funding and valid data. These problems are solved with the development of artificial general intelligence (AGI), especially LLMs such as the ChatGPT [4]. LLMs perform well at all stages of robot development and can be tasked to achieve collaborative work.

In this paper, we proposed a large-scale language model co-development framework to formulate automatic robot development strategies and generate precise executable code through the collaborative cooperation of multiple LLMs. To accomplish robot development more efficiently, we formulate the workflow as a team with development experience. We built a multi-tiered tasking system, in which multiple LLMs act as analysts, programmers, and testers. Analysts focus on analyzing user requirements in depth, providing detailed and accurate requirements guidance for code generation. Programmers are responsible for code creation, while testers make parameter adjustments based on user feedback to ensure that the code is applicable in the real-world environment of robot development. Each large-scale language model is tasked with solving a key complex problem in robot development. We utilize three ChatGPT agent models to play each of these three roles according to the role instructions and collaborate on robot development tasks, which are under the guidance of the large-scale language model co-development framework. To ensure the effectiveness of synergy among multiple models, we developed explicit collaboration rules to imitate the real-world robot development team collaboration patterns. Analysts, programmers, and testers work closely together and are responsible for key aspects of robot development such as strategy formulation, code generation, and parameter tuning [5].

The innovation of this automated collaboration framework is that individuals with no expertise in robotics can accomplish highly complex robot development tasks simply by interacting with the framework. Compared to the traditional multi-individual teamwork approach and the limitations of a single large-scale language model in handling complex tasks, our proposed approach demonstrates significant advantages in terms of efficiency and accuracy [6]. This research provides novel ideas and strategies for realizing automated robot development. Streamlining and embellishing this text

Our contributions include: (1) We propose a framework for collaborative development of large language models, where LLMs cooperate with each other to accomplish complex robot development tasks. (2) We formulate explicit collaboration rules that enable these models to work together as a real-world robotics development team. (3) This LLM co-development framework makes it possible for individuals with no expertise in the robotics domain to accomplish highly complex robot development tasks simply by interacting with the framework.

This paper is organized as follows: section 2 presents previous work related to LLMS inference. Section 3 describes our proposed methodology in detail. Section 4 describes the quadruped robot development experiments that we conducted and analyzes the results in depth. Finally, the conclusions of this study are drawn in Section 5.

## II. RELATED WORK

The development and applications of large language models (LLMs) have attracted substantial attention in various research domains. The following studies contribute to the understanding of LLMs' capabilities and their applications in different contexts. In the context of prompting methods, Wei et al. [7] introduced "chain-of-thought" prompting, which promotes reasoning in LLMs through interconnected prompts. Kojima et al. [8] explored the zero-shot reasoning potential of LLMs, revealing their ability to perform reasoning tasks without specific training data. Ouyang et al. [9] focused on training LLMs to follow instructions with human feedback, demonstrating the adaptability of LLMs in learning from interactions. Scaling instruction-finetuned language models gained attention, as exemplified by the work of Chung et al. [10]. Their research aimed to enhance the capabilities of instruction-based fine-tuning, contributing to LLMs' effectiveness in understanding and executing instructions. Program synthesis using LLMs was another significant area of exploration. Austin et al. [11] delved into program synthesis with LLMs, seeking to generate executable code from natural language prompts. Jain et al. [12] introduced "Jigsaw," a framework where LLMs meet program synthesis, enabling automatic code generation from high-level instructions. In the realm of multi-turn program synthesis, Nijkamp et al. [13] developed "Codegen" an open LLM model specializing in code generation through multi-turn interactions. Additionally, Chen et al. [14] focused on teaching LLMs to self-debug, enabling the models to identify and rectify errors in their generated code.

Furthermore, research has explored LLMs' interaction with software development. Jiang et al. [15] conducted a study on LLMs' abilities to improve the effectiveness of software evolution tasks. Shin et al. [16] introduced "AutoPrompt," a method for knowledge elicitation from LLMs using automatically generated prompts. Reynolds and McDonell [17] extended the concept of prompt programming beyond few-shot scenarios to enhance LLMs' expressiveness in generating code. Additionally, studies have examined strategies to assess LLMs' knowledge and understanding. Jiang et al. [18] proposed methodologies to ascertain the knowledge possessed by LLMs, shedding light on the reliability of LLMs' outputs. Collectively, these studies provide valuable insights into LLMs' reasoning abilities, prompting mechanisms, program synthesis applications, self-debugging capabilities, and interactions with software development tasks. The literature reviewed here enriches the understanding of LLMs and serves as a foundation for the collaborative development framework proposed in this paper.

## III. APPROACH

Our proposed framework for collaborative development of Large Language Models aims at an effective division of labor，where multiple Large Language Models play the roles of analysts, programmers, and testers, respectively, in robot development.

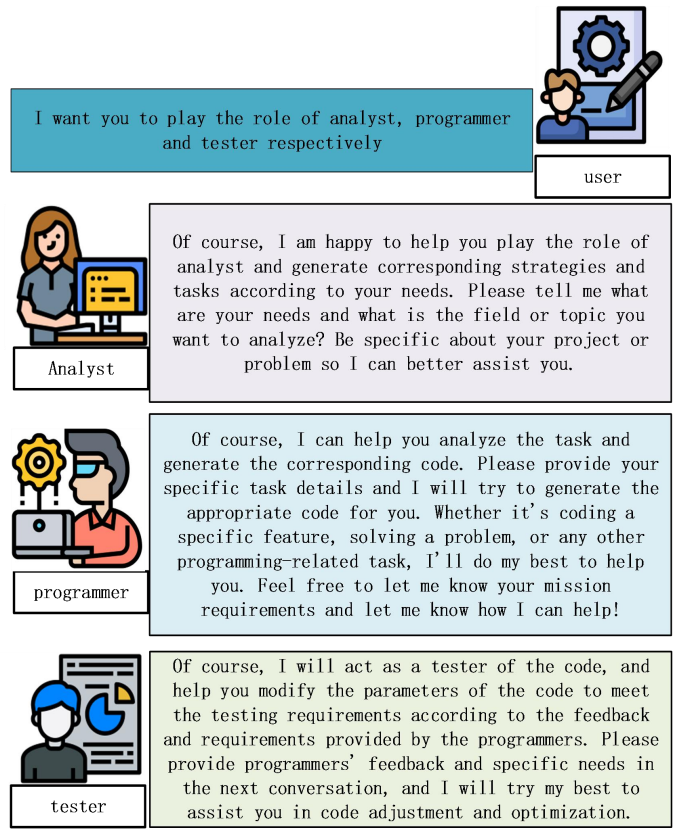

Fig. 1. LLMS role-playing in the robot development process.

By providing LLMs with specific instructions, we are able to direct their output based on user prompts, thus bringing the generated content closer to the intended solution. In our approach, to ensure that different LLMs are able to handle different subtasks, we enter different instructions for them, direct them to the appropriate roles, and specify the subtasks that need to be accomplished. Specifically, we require LLMs to participate in a special capacity related to their roles [19]. Also, we describe the roles we expect LLMs to play in the collaboration to simulate a real robot development team. In this way, LLMs were able to exhibit traits similar to professional robotics developers, with extensive development experience and expertise. As a result of our experiments, we found that the results generated by making LLMs play roles are more accurate compared to the results generated by asking LLMs direct questions, and therefore, assigning LLMs to different roles can significantly improve the ability to handle sub-tasks in the robot development process, as shown in Fig. 1.

We have clearly defined a series of sequential steps to ensure the smooth operation of the Large Language Model (LLM) collaborative development framework. These steps are described in detail below to ensure clear logic and rigor of the process: first, the collaboration starts with the analyst, who is responsible for receiving key information from the user, such as the nature of the task, goals, and constraints. Based on role instructions, the analyst translates this information into decisions that the programmer can understand, designs the decisions to facilitate subsequent operations, and provides clear guidelines for the programmer. The further breakdown of decisions into multiple specific subtasks is a key aspect of the entire collaborative process. This refinement process not only breaks down the task into manageable parts, but also helps programmers understand and operate more clearly. The programmer generates code based on the well-defined sub-tasks, ensuring that the code is consistent and accurate to the task. The generated code is passed to the testers. Testers play a crucial role in the collaboration, making necessary parameter adjustments based on robot feedback provided by human users. This critical step allows the code to meet the actual user requirements. If the requirements are not met after several adjustments, the tester passes the feedback to the analyst. This feedback loop introduces resilience and continuous improvement to the collaborative process, ensuring the quality of the generated code.

This is shown in Figure 2. This highly collaborative process allows LLM to work together under different roles, approximating the way real-world robotics development teams collaborate. In this way, we automate the collaboration of robot development to ensure task accuracy and efficiency. Based on the experimental results, the collaboration framework based on role division of labor significantly improves the quality and efficiency of robot development [20]. Through rational division of labor and orderly collaboration, we successfully direct LLM to different roles to provide accurate support and optimization for all stages of robot development.

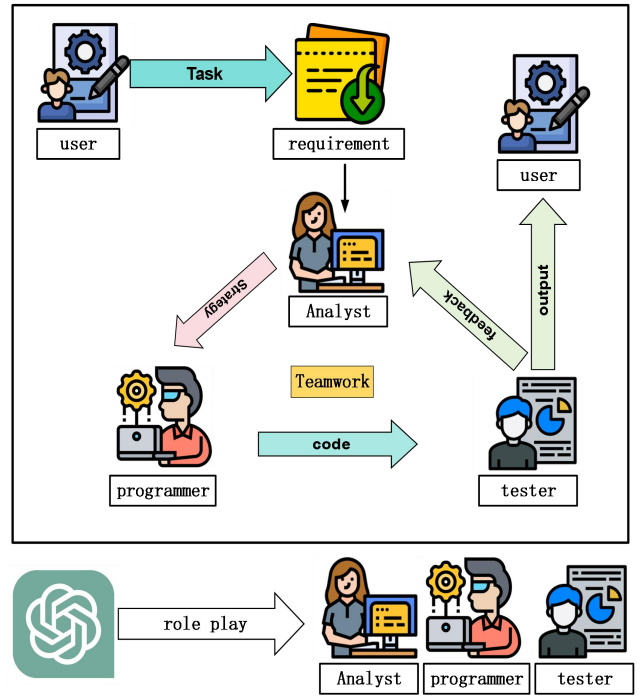

Fig. 2. Schematic diagram of LLMS collaboration for complex robot development

In this collaborative process shown in Fig 2, LLM works with each other in the roles of analyst, programmer, and tester, simulating the way real-world robot development teams work together. In this way, we not only automate the collaboration of the robot development process, but also ensure the accuracy and efficiency of the tasks. According to our experimental results, this collaborative framework based on the division of roles shows obvious advantages in improving the quality and efficiency of robot development.

## IV. EXPERIMENT

In this experiment, we chose a quadruped robot based on ROS (Robot Operating System) as shown in Fig. 3. We have developed a goal-following function on this robot, aiming to enable the robot to follow the movement of a worker and to perform obstacle avoidance and distance keeping functions. The main goal of our experiments is to validate the effectiveness of the proposed Large Language Model (LLM) co-development framework in this specific robot application scenario.

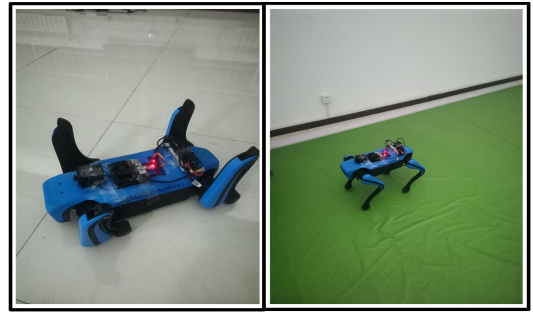

Fig. 3. The quadrupedal robot used for the experiment.

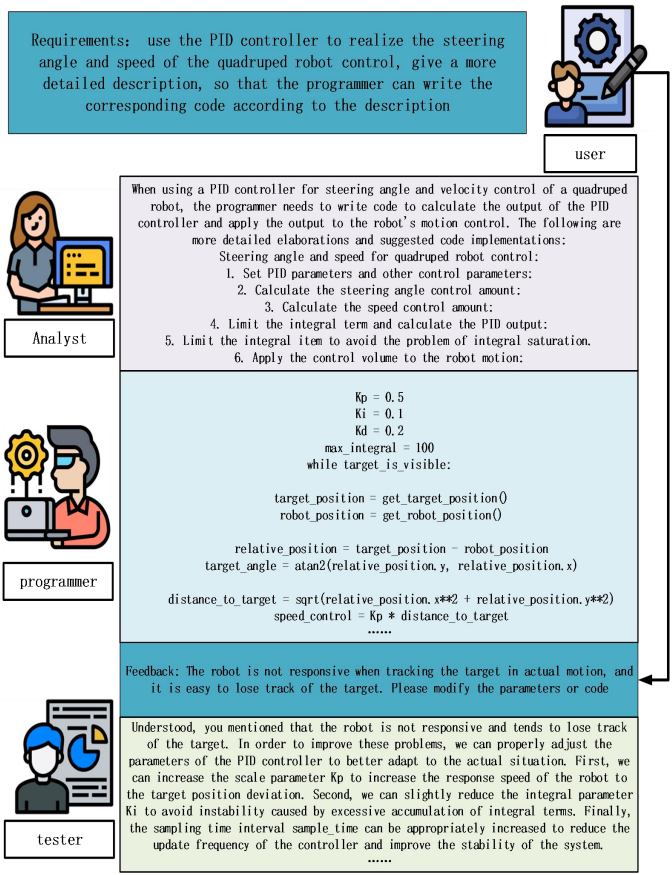

Fig. 4. Example of a large language model development robot

Figure 4 shows an example of the robot development process, where our experiments validate the effectiveness of LLM in the development of the quadruped robot's goal following function through a collaborative development framework [21]. Task decomposition and strategy generation by analysts and code generation by programmers, and the final code obtained was verified in real-world testing by testers. This experiment proves the practical application value of the proposed co-development framework in the field of robot development and provides a more efficient and precise method for robot development.

First, we provided the analyst with the equipment parameters and functional requirements for the quadruped robot. These parameters include the physical characteristics of the robot, the sensor configuration, and the motion control method. In addition, we clearly expressed the functional requirements for the robot development, which are to fulfill the tasks of mobile following, obstacle avoidance, and distance maintenance [22]. This information provided the analyst with critical background knowledge and guidance for task decomposition during the collaboration process. Next, the analysts were guided by role instructions to effectively decompose the robot development tasks. This included dividing the complex goal-following task into a series of more specific, actionable subtasks for programmers to understand and implement. The analyst generates strategies that are easy for the programmer to understand, specifying the behaviors and decisions that the robot needs to follow in order to achieve the desired functionality.

The programmer used LLM to generate appropriate code based on the strategies provided by the analyst. These codes include key functions such as movement control, obstacle avoidance and distance maintenance of the robot [23]. During the code generation process, LLM's role instructions ensured that the programmer was able to generate code that was consistent with the desired functionality based on an understanding of the analyst's strategy.

The generated code is then passed to a tester whose task is to evaluate the performance of the code on an actual robot. Using an actual quadruped robot platform, the tester simulates situations such as staff movement, obstacle avoidance, and distance maintenance to verify that the generated code meets the user's needs [24]. If needed, the tester makes appropriate parameter adjustments to the code based on the results of the experiments to ensure that the robot performs as expected in a real-world environment.

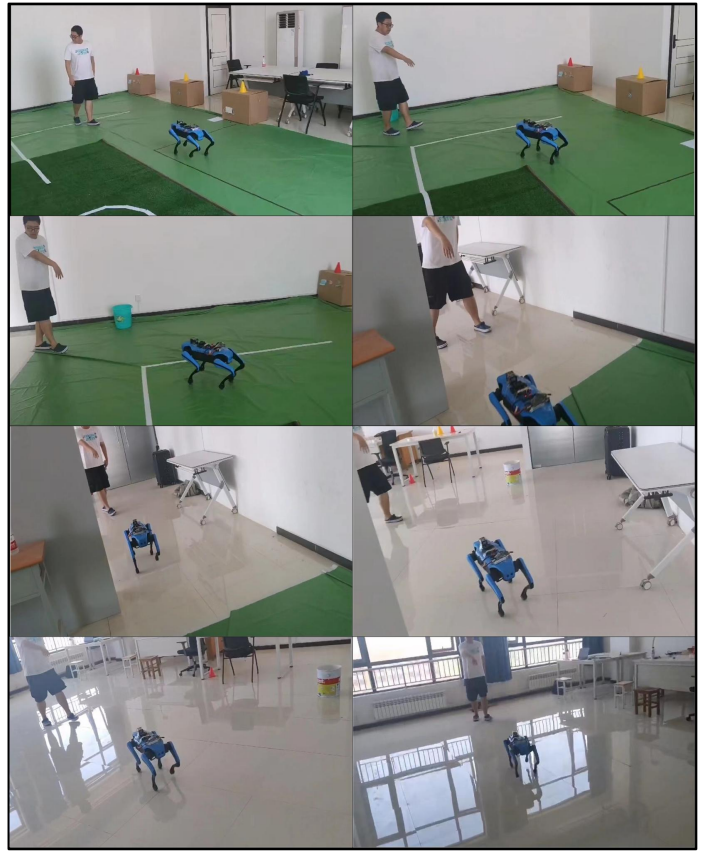

Fig. 5. Quadrupedal robot performs target tracking, following from one area to another.

As shown in Fig. 5, in the experimental phase, we successfully applied the co-development framework described above to realize the development of a quadrupedal robot that can effectively follow the movement of a person. Our experimental results not only confirm the feasibility and effectiveness of this approach in real robotics applications, but also provide a novel approach for automated development to cope with complex tasks.

Typically, realizing the people-following function of a quadruped-like robot requires team collaboration as well as a significant amount of time and resources. However, with our proposed co-development framework, this complex task can be successfully accomplished by only one person who has not been involved in development before. This result significantly highlights the superiority of the framework in lowering the development threshold and increasing the development efficiency.

It is worth emphasizing that a single Large Language Model (LLM) is not capable of developing such a complex robotics task. However, through the role division of labor and collaborative approach we presented, the LLM is subdivided into roles such as analyst, programmer, and tester, which achieves the goal of having each of them perform their respective roles and collaborate with each other. Using this model, we successfully accomplished the development of complex tasks for the robot. The results of this experiment undoubtedly confirm the excellence of the collaborative development framework in dealing with complex tasks.

Our experiments yielded promising results in real-world applications, successfully realizing the following function of a quadruped robot. This approach not only achieves a significant breakthrough in development efficiency, but also provides a new solution for complex tasks that cannot be handled by a single LLM model. Through this experiment, we not only verified the practicality of the co-development framework, but also opened up new possibilities for automated collaboration in the field of robot development, which effectively meets the practical needs of users.

## V. Conclusion

This paper presents a collaborative development framework based on Large Language Models (LLMs) that aims to address the challenges of complex robot development tasks in the real world. Although LLM has demonstrated amazing capabilities in code generation, there are still challenges in handling complex tasks. Robot development involves task assignment and teamwork, requiring the cooperation of specialized knowledge and extensive experience. Inspired by this, we propose an automated collaboration framework that puts LLMs in the roles of analysts, programmers, and testers to realize robot development tasks by simulating a realistic robot development team.

In the experimental phase, we apply the framework to the development of the goal-following function of a quadruped robot. The experimental results show that through the collaborative development framework, we successfully realized the complex functions of goal following, obstacle avoidance, and distance keeping for the quadruped robot. Compared with traditional methods, this framework not only reduces the development threshold, but also improves the development efficiency. It is difficult for a single LLM model to perform such a complex task, but through the division of roles and collaborative work, we have achieved efficient cooperation among the roles, thus fully utilizing the potential of LLM.

In summary, the LLM-based automatic collaboration framework proposed in this study brings new ideas and methods to the field of robot development. By assigning LLMs to different roles and simulating the cooperation of a robot development team, we successfully realized the automated development of complex robot tasks. The experimental results validate the feasibility and effectiveness of the framework in real-world applications and provide a new way to address the challenges in robot development. This research not only expands the application area of LLM, but also provides strong support for automated collaboration in robot development, which brings insights for future research and practice.

### Acknowledgment

This work is supported by the National Natural Science Foundation of China (No. U21A20485).